\documentclass[10pt,journal,compsoc]{IEEEtran}

\ifCLASSOPTIONcompsoc
  \usepackage[nocompress]{cite}
\else
  \usepackage{cite}
\fi
\usepackage{hyperref}
\usepackage{algorithm}
\usepackage{algorithmicx}
\usepackage{amsmath}
\usepackage{amssymb}

\usepackage{algpseudocode}
\usepackage{multirow} % peter add for table using
\usepackage{booktabs} % peter add for table
\usepackage{newfloat}
\usepackage{listings}

\usepackage{graphicx} % DO NOT CHANGE THIS
\usepackage{subfigure}

\usepackage{cite}
\ifCLASSINFOpdf
\else
\fi

\begin{document}
%\title{STDGRL: Spatio-Temporal Dynamic Graph Relation Learning for Urban Metro Flow Prediction}
\title{Spatio-Temporal Dynamic Graph Relation Learning for Urban Metro Flow Prediction}

\author{Peng Xie,
        Minbo Ma,
%        Fei Teng,% <-this % stops a space
        Tianrui Li, ~\IEEEmembership{Senior Member,~IEEE,} \protect \\
        Shenggong Ji, Shengdong Du, Zeng Yu 
    	 and Junbo Zhang, ~\IEEEmembership{Member,~IEEE,}
\IEEEcompsocitemizethanks{\IEEEcompsocthanksitem Peng Xie, Minbo Ma, Tianrui Li (the corresponding author), Shengdong Du, Zeng Yu are with School of Computing and Artificial Intelligence, Southwest Jiaotong University, Chengdu 611756, China. Tianrui Li is also with National Enginnering Laboratory of Integrated Transportation Big Data Application Technology.
E-mails: pengxie@my.swjtu.edu.cn, minboma@my.swjtu.edu.cn, trli@swjtu.edu.cn, sddu@swjtu.edu.cn, zyu@swjtu.edu.cn.
%\protect\\

\IEEEcompsocthanksitem Shenggong Ji is with Tencent Inc., Shenzhen, China. \protect\\
E-mail: shenggongji@163.com.

\IEEEcompsocthanksitem Junbo Zhang is with JD iCity, JD Technology, Beijing, China, JD Intelligent Cities Research $\&$ Institute of Artificial Intelligence, Southwest Jiaotong University, Chengdu 611756, China. \protect \\
E-mail: msjunbozhang@outlook.com. 
}% <-this % stops an unwanted space
\thanks{Manuscript received xxx xx, xxxx; revised xxxx xx, xxxx.}}
%\thanks{Manuscript received xxx 19, 2005; revised August 26, 2015.}}


\markboth{Journal of \LaTeX\ Class Files,~Vol.~14, No.~8, August~2015}%
{Shell \MakeLowercase{\textit{et al.}}: Bare Demo of IEEEtran.cls for Computer Society Journals}



\IEEEtitleabstractindextext{%
\begin{abstract}
Urban metro flow prediction is of great value for metro operation scheduling, passenger flow management and personal travel planning. However, it faces two main challenges. First, different metro stations, e.g. transfer stations and non-transfer stations, have unique traffic patterns. Second, it is challenging to model complex spatio-temporal dynamic relation of metro stations. To address these challenges, we develop a spatio-temporal dynamic graph relational learning model (STDGRL) to predict urban metro station flow. First, we propose a spatio-temporal node embedding representation module to capture the traffic patterns of different stations. Second, we employ a dynamic graph relationship learning module to learn dynamic spatial relationships between metro stations without a predefined graph adjacency matrix. Finally, we provide a transformer-based long-term relationship prediction module for long-term metro flow prediction. Extensive experiments are conducted based on metro data in Beijing, Shanghai, Chongqing and Hangzhou. Experimental results show the advantages of our method beyond 11 baselines for urban metro flow prediction.
\end{abstract}

}

\maketitle

\IEEEdisplaynontitleabstractindextext

% \fi
\IEEEpeerreviewmaketitle


\section{Introduction}\label{sec:introduction}

\begin{figure}
	\centering
	\includegraphics[width=0.5\textwidth]{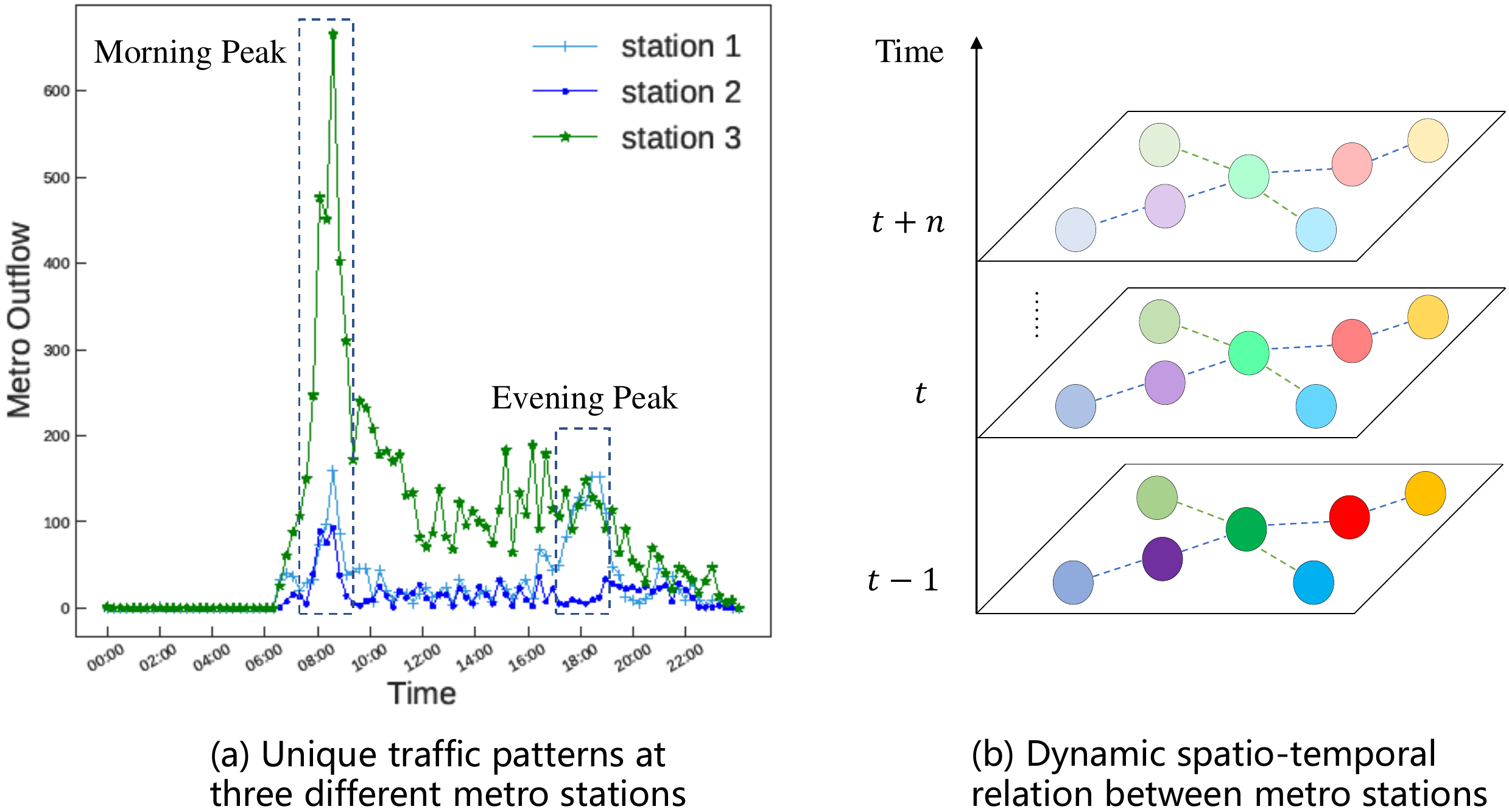}
	\caption{Spatio-temporal dynamic graph relation. (a) The metro outflow of station 1 has a prominent morning peak and evening peak. As a contrast, station 2 only has a morning peak and station 3 has an extremely big morning peak. It shows unique traffic patterns at three different metro stations. (b) These stations are connected to each other, the passengers' inflow and outflow of metro stations change over time, there are dynamic spatio-temporal dependency relations between stations.} \label{fig:DSG-v4}
\end{figure}

\IEEEPARstart{A}{s} a significant part of urban public transportation, the urban metro takes a large proportion of urban traffic. Especially for large cities, accurate prediction of urban metro passenger flow is significant for metro operation scheduling \cite{gong2020online}, passenger flow management \cite{gong2020potential}, and personal travel planning \cite{cheng2020analysis}. The urban metro network is a spatio-temporal dynamic graph with prominent spatial and temporal characteristics. We show the change of passenger outflow for three different metro stations over the time frame of one day in Figure \ref{fig:DSG-v4}(a). We can observe  that the passenger outflow of station 1 has a small peak between 7:00 and 9:00 in the morning, and there is also a small evening peak period between 17:00 and 19:00. While station 2 only has a relatively small period of passenger flow between 7:00 and 9:00 in the morning, there is no obvious evening peak at night, and the overall one-day passenger outflow is smaller than that of station 1. Station 3 has a large peak in passenger outflow between 7:00 and 9:00, and then the passenger outflow at other periods decreases significantly. Still, the overall passenger flow of station 3 is much larger than those of stations 1 and 2. We can see that these stations have their own different station traffic patterns, not just a simple, fixed spatial connection relationship between stations. Different metro stations are connected and affected each other. This spatial dependency relationship changes dynamically along with time and location as shown in Figure \ref{fig:DSG-v4}(b).

In order to achieve the prediction of passenger flow in metro stations, some research works have been tried and studied \cite{tang2018forecasting, liu2019deeppf, zhang2020deep, chen2021graph}. Most of these methods model the flow change trend of metro stations according to inflow and outflow passenger data, metro network topology map, weather, and other external factors. Most of them use CNN and GNN-based methods to capture spatial dependencies in metro flow data \cite{ye2020multi}, and apply RNN-based and Attention-based methods to model the temporal dependencies of metro traffic data \cite{ma2018parallel}, and some research take external factors into account \cite{zhang2020deep}. These studies have made some progress, but most of these works only use a single metro traffic data set or need to predefine the adjacency relationship between stations. Others treat different stations in the metro network as the same kind of node. Overall, the generalization performance of these models is insufficient. 

%Our research does not require a predefined station graph adjacency matrix compared to previous work. And it learns the spatial dependency relationship between stations dynamically and treats different stations as graph nodes with varying traffic patterns. Experiments are carried out on multiple metro traffic datasets to verify the generalization performance and scalability of the model.

In summary, the urban metro flow prediction task faces three major challenges:

1) \textbf{Modeling unique traffic patterns at different stations:} Previous research \cite{ye2020multi, ou2020stp, liu2020physical} has treated metro stations as equal nodes or divided metro stations into transfer stations and non-transfer stations. The parameters are shared globally or locally when using a static adjacency matrix, and the computational cost is relatively small. Still, it ignores the traffic flow patterns differences between different stations. However, we find that although different stations are directly connected or are all transfer stations, they have unique traffic change patterns, as shown in Figure \ref{fig:DSG-v4}(a). Therefore, it is necessary to model the traffic patterns of different stations separately.

2) \textbf{Dynamic spatial dependency relations between stations:} The spatial dependencies between stations are regarded as static in the existing work \cite{ye2020multi, chen2020physical, zhang2020deep}. Some of them express their spatial dependencies directly with the existence or lack of connections between stations. The distance between them and the similarity of traffic flow is regarded as spatial dependencies. But this method is viewed as static, which ignores that the passenger inflow and outflow of a station are not only affected by the passenger flow of its upstream, downstream stations and nearby stations, but also by time, weather and other factors. Therefore, it's a challenge to capture the dynamic spatial dependency relation between stations.

3) \textbf{Long-term temporal prediction:} To better support the downstream applications of the metro, it is necessary to carry out a long-term metro station flow prediction. Existing research \cite{zhang2020deep} on short-term metro station passenger flow prediction has been carried out. Still, there is a lack of relevant research on long-term accurate metro station flow prediction because long-term time series forecasting is challenging. As the prediction period becomes longer, the influence of uncertain factors will reduce prediction accuracy, and the dynamic variability of the metro flow itself also increases the uncertainty of metro flow prediction. In general, compared with short-term prediction, long-term prediction is more difficult but has greater practical application value.

In order to cope with the above challenges, we propose a spatio-temporal dynamic graph relation learning method for metro flow prediction, which can model different traffic patterns at different stations and capture the dynamic spatial dependency relation between stations. At the same time, it can carry out long-term prediction, which can better support traffic management of metro operation managers and travel decisions of urban residents. The contributions of this paper include four aspects, as follows:

\begin{itemize}
	
	\item A node-adaptive parameter learning module is adopted to learn different station-specific spatiotemporal embedding representations to capture the flow patterns of different stations.
	
	\item A dynamic graph relation learning module is proposed to learn the dynamic spatial dependencies between stations, which does not require a predefined spatial relationship of station connections, and directly learn the dynamic spatial dependencies between stations from the spatiotemporal graph data.
	
	\item A long-term temporal relation prediction module based on Transformer is used to predict the long-term metro flow. The predicted results can offer a useful reference for urban metro operation management and personal travel planning.
	
	\item Experiments are conducted on 4 different cities' metro datasets, including Beijing, Shanghai, Chongqing, Hangzhou. Compared with the 11 baseline methods, the experimental results have significantly improved prediction performance.
	
\end{itemize}

The remainder of this paper is organized as follows. In Section 2, we present the related work about urban flow prediction and graph neural networks. In Section 3, we introduce some preliminary concepts and formalize the metro flow prediction problem. In Section 4, we show the overall framework of the proposed STDGRL model. The experiment result, visualization and analysis are given in Section 5. We conclude the work in Section 6.

\section{Related Work} \label{section: works}

\subsection{Urban Flow Prediction}
Urban flow prediction is important for traffic management\cite{chen2016learning}, land use\cite{jayarajah2018understanding}, public safety\cite{zhang2017deep}, etc. The urban flow prediction can be regarded as a spatio-temporal prediction task, which is a kind of research problem that uses spatio-temporal machine learning methods to learn spatio-temporal correlations from spatio-temporal datasets\cite{DBLP:conf/gis/ZhangZQLY16}. At present, a large number of researchers have conducted studies on the task of urban flow prediction. Xie et al.\cite{DBLP:journals/inffus/XieLLDYZ20} divided the urban flow prediction task into crowd flow prediction, traffic flow prediction, and public transport flow prediction and reviewed the classical deep learning methods. With the city's continuous development, more and more people are pouring into the city, and the metro and other public transportations occupy the main body of the urban traffic flow. Accurate metro flow prediction is of great value for urban traffic management, urban public safety, and residents' daily travel. In the early work, researchers used statistical-based methods for urban flow prediction, such as ARIMA (Autoregressive Integrated Moving Average)\cite{williams1998urban}, SARIMA (Seasonal Auto-Regressive Integrated Moving Average)\cite{zhang2011seasonal} and other methods. Later, some classic machine learning methods were used for urban flow prediction, such as SVR(Support Vector Regression)\cite{lippi2013short}, K-NN(K-nearest neighbor)\cite{habtemichael2016short} and other methods. But these methods often ignored spatiotemporal correlations are hinted in spatiotemporal data, which are crucial for accurate urban flow prediction.

In recent years, with the development of deep learning, deep learning methods have been used in the research field of urban flow prediction. The representative works mainly include the time series method represented by RNN\cite{zhao2017lstm}, the spatial relation method represented by CNN\cite{jiang2018geospatial}, and a spatiotemporal relationship method combining the two\cite{wu2018hybrid,duan2018improved,ma2018parallel}. Based on RNN and its variant series, these methods focus on capturing temporal dependencies in spatio-temporal data, such as closeness, periodicity, trend, etc\cite{zhang2017deep}. These CNN-based methods mainly capture the spatial dependencies in spatiotemporal data, such as spatial distance, spatial hierarchy, and regional functional similarity\cite{zhang2019flow}. In addition, such methods combining RNN and CNN consider both temporal and spatial dependencies and propose hybrid models to model the spatiotemporal characteristics in traffic data\cite{yao2019revisiting}.

Later, due to the rise and continuous development of the graph neural network\cite{bruna2014spectral,defferrard2016convolutional,DBLP:conf/iclr/KipfW17} and the graph structure of the road network and rail transit network, more and more researchers have used GNN-based methods for urban flow prediction tasks\cite{DBLP:journals/corr/abs-2101-11174,DBLP:conf/iclr/LiYS018,yu2018spatio} and achieved good results.

\subsection{Graph Neural Networks}
Graph neural networks can model graph data in non-Euclidean space, especially the dependencies between nodes. Wu et al. \cite{wu2020comprehensive} divided graph neural network methods into graph convolutional networks, graph attention networks, graph autoencoders, graph generation networks, and graph spatiotemporal networks. Applying the graph neural network to urban flow prediction, traffic forecasting, and other fields is natural. Since the road network and rail transit network can be regarded as the road segments and stations in the graph, the graph spatiotemporal network can be used to capture the relationship between the nodes. Based on RNN and CNN, the spatial and temporal dependencies in the spatio-temporal graph can be learned, making more accurate traffic state predictions. Among them, two representative works use GCN and RNN\cite{DBLP:conf/iclr/LiYS018}, GCN and CNN\cite{yu2018spatio} methods to model the spatiotemporal dependencies of spatiotemporal graph data, which are applied to traffic prediction tasks.

However, the previous methods using GNNs for spatiotemporal prediction tasks mostly use a predefined graph structure or a single fixed graph adjacency matrix\cite{zhao2019t} or multiple graph adjacency matrices for fusion\cite{chen2020physical}. This type of method regards the spatial dependence in spatiotemporal data as static and invariant. However, in reality, the spatiotemporal relationship in spatio-temporal data is dynamic. It is necessary to model the dynamic graph relationship in spatio-temporal data and capture the spatio-temporal dynamics. Compared with previous methods, our method mainly learns the dynamic graph relationship in the spatiotemporal data to obtain more accurate traffic prediction results.

\section{Problem Formulation}\label{section: formulation}
This paper proposes a spatio-temporal dynamic graph relation learning model for flow prediction in metro stations. Our model does not need a predetermined metro network topology map, and can directly learn spatial dependencies from metro flow data, which has broad applicability to metro flow prediction tasks in different cities.

Before introducing our model in detail, we first define and represent the metro flow prediction task and related conceptual notations. At station $i$, the metro flow of time period $t$ can be expressed as $\boldsymbol{X}_{i, t} \in \mathbb{R}^{2}$, which includes the passenger inflow and outflow. The flow information of the entire metro network can be expressed as $\boldsymbol{X}_{:, t} = (\boldsymbol{X}_{1, t}, \boldsymbol{X}_{2, t}, ... , \boldsymbol{X}_{N, t}) \in \mathbb{R}^{N \times 2}$, where $N$ means the number of metro stations. The metro flow in this paper contains two perspectives, which are passengers inflow and outflow in metro stations. The metro station flow prediction task can be defined as, given the historical flow sequence, predicting the flow sequence for a period of time in the future.

\begin{equation}
	X_{:, t+1}, X_{:, t+2}, \ldots, X_{:, t+m}=\mathcal{F}_{\theta}\left(X_{:, t}, X_{:, t-1}, \ldots, X_{:, t-T+1}\right),
\end{equation}
where $\theta$ means all the learnable parameters in the STDGRL model, $T$ is the length of the input flow sequence, and $m$ means the length of the predicted flow sequence.

\section{Methodology}\label{section: method}
\begin{figure*}
	\centering
	\includegraphics[width=1.0\textwidth]{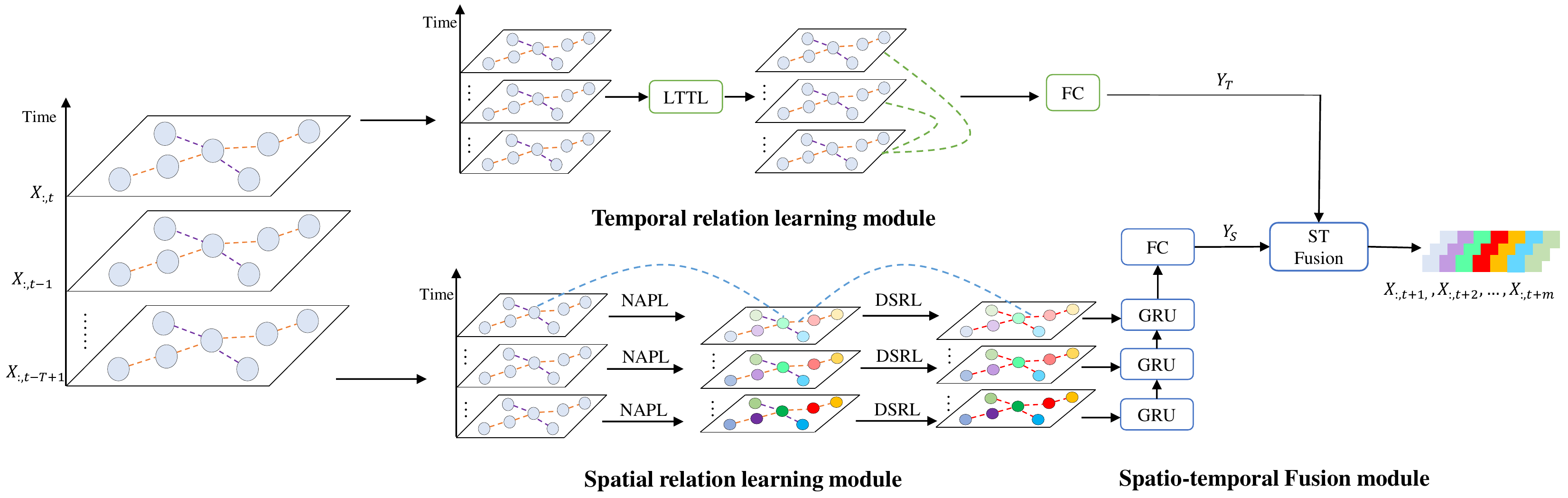}
	\caption{Spatio-Temporal Dynamic Graph Relation Learning (STDGRL) model.} \label{fig:Fig_STDGRL-v6}
\end{figure*}

The overall architecture of the model is shown in Figure \ref{fig:Fig_STDGRL-v6}. It contains a node-specific spatiotemporal embedding module, a dynamic spatial relationship learning module, a long-term temporal prediction module and a spatio-temporal fusion module. First, we propose a node-specific spatio-temporal embedding module to embed and represent the stations of the metro spatio-temporal graph. Then we adopt a dynamic spatial relationship learning module to learn the spatial dependencies directly from the metro flow data without relying on a specific metro network topology. Finally, a Transformer-based long-time-series dependency prediction module is used to predict the metro flow in a long-term sequence, making its prediction more suitable for actual metro dispatch management and daily operation scenarios.

\subsection{Node-specific Spatio-Temporal Embedding}
The node-specific adaptive parameter learning module (NAPL) is adopted. The classic graph convolution operation \cite{DBLP:conf/iclr/KipfW17} is calculated by the following formula:

\begin{equation}
	Z=\left(I_{N}+D^{-\frac{1}{2}} A D^{-\frac{1}{2}}\right) X \Theta+b,
	\label{equ:gcn}
\end{equation}
where $\boldsymbol{A} \in R^{N \times N}$ is the adjacency matrix of the graph, $D$ is the degree matrix, $I_{N}$ is the identity matrix, $\mathrm{X} \in R^{N \times C}$ is the input of the graph convolutional network layer, $Z \in R^{N \times F}$ is the output of the graph convolutional network layer, $\Theta \in R^{C \times F}$ and $\mathbf{b} \in R^{F}$ represent learnable weights and biases, respectively.

In this method, all nodes on the graph share parameters such as weights and biases. According to the viewpoint put forward by \cite{bai2020adaptive}, different nodes have different traffic flow patterns, as shown in Figure \ref{fig:DSG-v4}(a), because different nodes have different attributes, such as POI distribution around the nodes, various weather conditions, and different flow patterns will be formed. For more accurate traffic prediction, it is necessary to learn different traffic patterns for different nodes, that is, to learn node-specific patterns by using different learnable parameters rather than globally shared parameters.

In order to learn node-specific patterns, a node-specific adaptive parameter learning module is proposed, which learns the node embedding matrix $E_{\mathcal{G}} \in R^{N \times d}$ and weight pool $W_{\mathcal{G}} \in R^{d \times C \times F}$. The $\Theta$ in the formula \ref{equ:gcn} can be calculated by the node embedding matrix and the weight pool, $\Theta=E_{\mathcal{G}} \cdot W_{\mathcal{G}}$. Such a computation can be interpreted as learning node-specific patterns from all station time-series patterns. The bias $b$ can also be calculated in the same way. The parameter module of the final node adaptation can be expressed by the formula \ref{NAPL-GCN}.

\begin{equation}
	Z=\left(I_{N}+D^{-\frac{1}{2}} A D^{-\frac{1}{2}}\right) X E_{\mathcal{G}} W_{\mathcal{G}}+E_{\mathcal{G}} \mathrm{b}_{\mathcal{G}}\label{NAPL-GCN}.
\end{equation}

\subsection{Dynamic Spatial Relation Learning}
\begin{figure}
	\centering
	\includegraphics[width=0.40\textwidth]{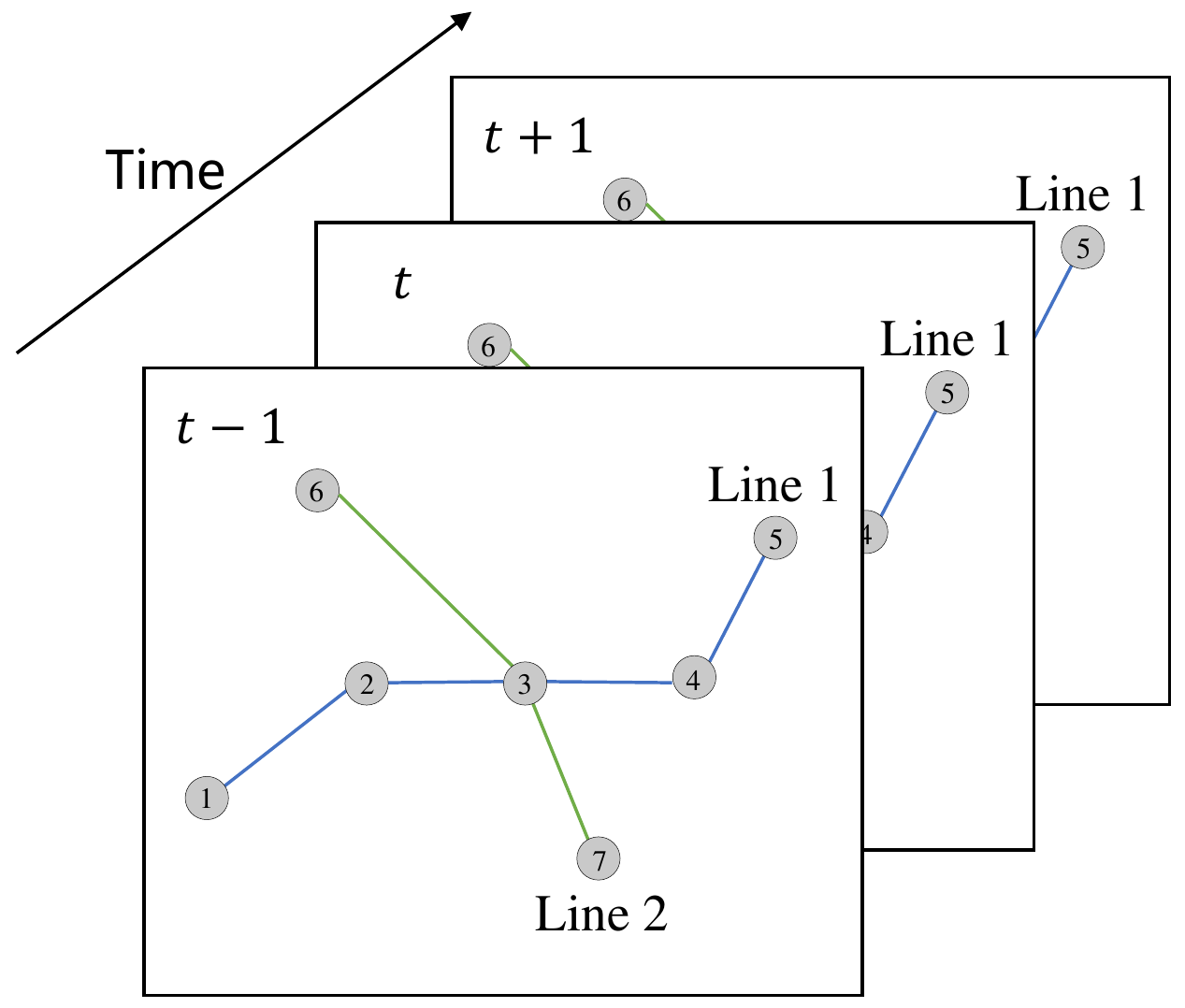}
	\caption{Static spatial relation between metro stations.} \label{fig:Fig_static_graph}
\end{figure}

In the metro network, the connection relationship between stations is fixed and static, as shown in Figure \ref{fig:Fig_static_graph}. However, this fixed and static connection relationship cannot reflect the dynamic spatial dependence between stations, and with the variation of time, the passengers' inflow and outflow of stations change, so it is necessary to learn this dynamic spatial dependency from spatiotemporal data. Therefore, a dynamic spatial relationship learning module (DSRL) is proposed, a representation model with adaptive and spatial structure awareness. Inspired by \cite{bai2020adaptive}, we first randomly initialize a learnable node embedding dictionary $E_A \in R^{N \times d_e}$ for all nodes. During the model training process, $E_A$ will be dynamically updated. Each row of $E_A$ represents the embedding representation of the node, and $d_e$ represents the dimension of node embedding. Then, the spatial dependency between nodes is calculated by multiplying $E_A$ and $E_A^T$. Finally, we can get the generated graph Laplacian matrix as shown in the formula below.

\begin{equation}
	D^{-\frac{1}{2}} A D^{-\frac{1}{2}} = softmax(ReLU(E_A \cdot E_A^T)),
\end{equation}
where the $softmax$ function is used to normalize the learned adaptive matrix. The calculation formula of GCN is as follows:
\begin{equation}
	Z = (I_N + softmax(ReLU(E_A \cdot E_A^T))) X \Theta + b. \label{equ:dsrl}
\end{equation}

For the nodes at time step $t$, the operation of the GRU module can be expressed as follows:

\begin{equation}
	\begin{aligned}
		&\widetilde{A}=\operatorname{softmax}\left(\operatorname{ReLU}\left(\boldsymbol{E_A} \boldsymbol{E_A^{\boldsymbol{T}}}\right)\right), \\
		&z_{t}=\sigma_z\left(\widetilde{\boldsymbol{A}}\left[\boldsymbol{X}_{:, t}, h_{t-1}\right] \boldsymbol{E} \boldsymbol{W}_{\boldsymbol{z}}+\boldsymbol{E} \boldsymbol{b}_{\boldsymbol{z}})\right., \\
		&r_{t}=\sigma_r\left(\widetilde{\boldsymbol{A}}\left[\boldsymbol{X}_{:, t}, h_{t-1}\right] \boldsymbol{E} \boldsymbol{W}_{\boldsymbol{r}}+\boldsymbol{E} b_{r})\right., \\
		&\hat{h}_{t}=\tanh \left(\widetilde{\boldsymbol{A}}\left[\boldsymbol{X}_{:, t}, r \odot \boldsymbol{h}_{t-1}\right] \boldsymbol{E} \boldsymbol{W}_{\hat{h}}+\boldsymbol{E} b_{\hat{h}})\right., \\
		&h_{t} \equiv \boldsymbol{z} \odot h_{t-1}+(1-\boldsymbol{z}) \odot \hat{h}_{t},
	\end{aligned}
\end{equation}
where $[\cdot]$ means the concate operation, $\odot$ denotes the element-wise multiplication, $\boldsymbol{E}$, $\boldsymbol{W}_{\boldsymbol{z}}$, $\boldsymbol{W}_{\boldsymbol{r}}$, $\boldsymbol{W}_{\hat{h}}$, $\boldsymbol{b}_{\boldsymbol{z}}$, $b_{r}$, $b_{\hat{h}}$ are the parameters to be learned, $X_{:, t}$ and $h_{t}$ are input and output at time step $t$. Finally, the output $Y_S$ of the component is obtained through a fully connected network.

\subsection{Long Term Temporal Prediction}
To capture the long-term global dependencies of metro flow sequences, we propose a long-term Transformer layer (LTTL). A Transformer-based\cite{DBLP:conf/nips/VaswaniSPUJGKP17} long-term temporal prediction method is adopted for long term metro flow  prediction. This layer includes a multi-head self-attention layer, a feed-forward neural network layer, and a layer normalization layer. First, the multi-head self-attention layer is introduced. The attention calculation formula is shown in the formula \ref{attention}. The dot product between all keys and the given queries is calculated, divided by $\sqrt{d_{k} }$, and then multiplied by $V$. Finally, a softmax function is used to calculate the attention score of each position. These attention scores will be used as weights to aggregate information from different parts.

\begin{equation}
	\text { Attention }(\mathrm{Q}, \mathrm{K}, \mathrm{V})=\operatorname{softmax}\left(\frac{\mathrm{QK}^{T}}{\sqrt{d_{k}}} \mathrm{~V}\right)\label{attention},
\end{equation}
where $Q, \mathrm{~K} \in \mathbb{R}^{T \times d_{k}}$ and $\mathrm{V} \in \mathbb{R}^{T \times d_{ v}}$ mean the queries, keys and values of all nodes, respectively. A position embedding is added to each position to enable the LTTL layer to perceive the relative position in the entire traffic sequence. The formula of position coding $e_{t}$ is shown below:

\begin{equation}
	e_{t}= \begin{cases}\sin \left(t / 10000^{2 i / d_{\text {model }}}\right), & \text { if } t=0,2,4 \ldots \\ \cos \left(t / 10000^{2 i / d_{\text {model }}}\right), & \text { otherwise.}\end{cases}
\end{equation}

Then, the output calculated by the multi-head self-attention layer is passed to the feedforward neural network layer. Finally, the output $Y_T$ of the LTTL network is obtained through the residual connection \cite{DBLP:conf/cvpr/HeZRS16} and layer normalization.

\subsubsection{Spatio-temporal Fusion}
In order to effectively utilize the captured temporal and spatial dependencies, we adopt the spatio-temporal fusion module to fuse the learned temporal and spatial dependencies. As shown in the following formula:

\begin{equation}
	X_{:, t+1}, X_{:, t+2}, \ldots, X_{:, t+m}=\mathbf{W}_{S} \odot \mathbf{Y}_{S}+\mathbf{W}_{T} \odot \mathbf{Y}_{T},
\end{equation}
where $\mathbf{Y}_{S}$ is the output of spatial relation learning module, 
$\mathbf{Y}_{T}$ is the output of temporal relation learning module, $\odot$ is the Hadamard product, $\mathbf{W}_{S}$ and $\mathbf{W}_{T}$ are the learnable weight parameters.

\section{Experiments}\label{section: experiment}
In this section, we first introduce the experimental setup, including the description of the dataset, experimental environment, implementation details, and evaluation metrics. Next, we compare our proposed method STDGRL with 11 representative methods. Finally, we conduct extensive experiments and analyze the effectiveness of our model and each module.

\subsection{Experiments Settings}
1) \textbf{Dataset description}: In this paper, we use 1 private metro card swiping dataset: Chongqing Metro dataset and 3 public metro card swiping datasets: Shanghai Metro dataset\cite{chen2020physical}, Hangzhou Metro dataset\cite{chen2020physical} and Beijing Metro dataset\cite{zhang2020deep}. The descriptive information about the four datasets is shown in Table \ref{tab:chap:table_1}.

CQMetro: This dataset is obtained by preprocessing the Chongqing metro swiping card data. We divide the data into 15-minute time slices to get the inflow and outflow passengers of the stations within the time slice. The time span is from March 1 to March 31, 2019. The Chongqing Metro dataset contains a total of 170 stations. The training set, validation set, and test set are divided in a chronological order according to the ratio of 6:2:2.

SHMetro: This dataset uses the Shanghai Metro dataset published in \cite{chen2020physical}, and the format of the dataset is consistent with the original paper. The time slice size is 15 minutes, and the time span is from July 1 to September 30, 2016. The Shanghai Metro dataset contains a total of 288 stations. The dataset is divided into training, validation, and test sets. The time range of the training set is from July 1 to August 31, 2016, and the time range of the validation set is from September 1 to September 9, 2016. The time frame of the test set is from September 10 to September 30, 2016.

HZMetro: This dataset also uses the Hangzhou Metro dataset published in \cite{chen2020physical}. The format of the dataset is consistent with the original paper. The time slice size is 15 minutes, and it contains 80 stations. The time frame is January 2019, with a total of 25 days. The time range of the training set is from January 1 to January 18, 2019, the time range of the validation set is from January 19 to January 20, 2019, and the time range of the test set is January 21 to January 25, 2019.

BJMetro: This dataset collects the data of Beijing Metro for five consecutive weeks from February 29 to April 3, 2016. It contains 17 metro lines and 276 metro stations, excluding the Airport Express and its stations.

2) \textbf{Implementation details}: We use the deep learning framework PyTorch\cite{paszke2019pytorch} to implement the model STDGRL in this paper and the deep learning models in the comparison methods. The experimental equipment uses a GPU card with an NVIDIA Titan V. In the Chongqing Metro data set, the card swiping data between 23:00-06:30 every day is directly deleted. Since this period is not within the operating time range of the metro, no passengers enter or leave the stations. We normalized the dataset in the same way as used in AGCRN\cite{bai2020adaptive}. The Adam\cite{kingma2014adam} optimizer is used to optimize our model. We take the data of the 12 historical time steps as input and the data of the next 12 time steps as output. Although our proposed method does not require a predefined adjacency matrix graph, we use the predefined adjacency matrix graph method as a contrasting method.

3) \textbf{Evaluation metrics}: We use three metrics commonly used in spatiotemporal prediction tasks, Mean Absolute Error (MAE), Root Mean Square Error (RMSE), and Mean Absolute Percentage Error (MAPE), to evaluate the performance of the method. The formulae are as follows:

\begin{itemize}
	\item Mean Absolute Error (MAE)
	\begin{equation}
		M A E=\frac{1}{n} \sum_{i=1}^{n}\left|\hat{y}_{i}-y_{i}\right|.
	\end{equation}
	
	\item Root Mean Square Error (RMSE)
	\begin{equation}
		R M S E=\sqrt{\frac{1}{n} \sum_{i=1}^{n}\left(\hat{y}_{i}-y_{i}\right)^{2}}.
	\end{equation}
	
	\item Mean Absolute Percentage Error (MAPE)
	\begin{equation}
		M A P E=\frac{100 \%}{n} \sum_{i=1}^{n}\left|\frac{\hat{y}_{i}-y_{i}}{y_{i}}\right|,
	\end{equation}
	
\end{itemize}
where $n$ is the number of test samples, $\hat{y}_{i}$ and $y_{i}$ mean the predicted passenger flow and the actual passenger flow, respectively. $\hat{y}_{i}$ and $y_{i}$ are transformed into the scale of the original value by inverse Z-score normalization.

\begin{table*}[ht]
	\centering
	\caption {Four cities metro datasets.}
	\label{tab:chap:table_1}
	\begin{tabular}[c]{c|c|c|c|c}
		\toprule
		{Dataset} & {CQMetro} & {SHMetro} & {HZMetro} & {BJMetro}\\
		\midrule
		City & Chongqing, China & Shanghai, China & Hangzhou, China & Beijing, China\\
		Station & 170 & 288 & 80 & 276\\
		Time Interval & 15min & 15min & 15min & 15min\\
		Time Span & 3/1/2019-3/31/2019 & 7/1/2016-9/30/2016 & 1/1/2019-1/25/2019 & 2/29/2016-4/3/2016\\
		\bottomrule
	\end{tabular}
\end{table*}

\subsubsection{Baselines}
In this section, we compare the proposed STDGRL model with 11 baseline models. These models can be divided into five categories, including (1) two traditional time series models, (2) two single deep learning models, (3) five graph spatiotemporal network models for traffic prediction proposed in recent years, (4) one Transformer-based traffic prediction model, and (5) one recently proposed graph neural network model for metro passenger flow prediction. These models are described in detail as follows:

\begin{itemize}
	\item \textbf{Historical Average (HA)}\cite{liu2004summary}: This model obtains the current traffic by averaging the historical traffic in the same time slice. This method is calculated for a single time series each time.
	
	\item \textbf{Support Vector Regression (SVR)}\cite{smola2004tutorial}: This machine learning model serves as a classic baseline model for a class of time series forecasting, using linear support vector machines for time series forecasting tasks. It is often used as a comparison method in time series forecasting tasks.
	
	\item \textbf{Long Short-Term Memory (LSTM)}\cite{hochreiter1997long}: This is a classic deep learning method for time series that captures the temporal correlations of spatiotemporal sequences.
	
	\item \textbf{Gated Recurrent Unit (GRU)}\cite{cho2014properties}: As a variant model of RNN, it can also capture the time-series correlation in the spatiotemporal sequence, but it cannot learn the spatial correlation. It is a time series forecasting method based on deep learning.
	%	\item \textbf{Graph Convolutional Network (GCN)}\cite{kipf2016semi}: Graph Convolutional Networks are used to capture the spatial dependencies of nodes' neighbors as a contrasting model based on graph neural networks.
	
	\item \textbf{T-GCN}\cite{zhao2019t}: It is a traffic prediction model based on graph convolutional network, which can capture spatiotemporal dependencies in spatiotemporal sequence data. It combines a graph convolutional neural network and a gated recurrent neural network.
	
	\item \textbf{DCRNN}\cite{DBLP:conf/iclr/LiYS018}: To capture the complex spatial dependencies and nonlinear temporal dynamics of road networks, a diffusion convolutional recurrent neural network is proposed for traffic prediction. It is one of the classic methods for spatiotemporal sequence prediction in graph neural network-based methods.
	
	\item \textbf{STGCN}\cite{yu2018spatio}: This is a spatiotemporal graph convolutional network based on convolutional structure, and it is used for the traffic prediction task. It has a faster training speed and fewer parameters.
	
	\item \textbf{AGCRN}\cite{bai2020adaptive}: This method does not require a predefined spatial graph and is an adaptive graph convolutional network that can learn spatiotemporal dependencies from spatiotemporal data.
	
	\item \textbf{Graph WaveNet}\cite{wu2019graph}: It uses a node embedding method to learn the adaptive spatial graph structure, a spatiotemporal graph network method combining graph convolution and dilated causal convolution is proposed.
	
	\item \textbf{STTN}\cite{xu2020spatial}: It's a Transformer-base spatio-temporal model for traffic prediction.
	
	\item \textbf{Multi-STGCnet}\cite{ye2020multi}: It is a combined model containing graph convolutional network and LSTM for metro passenger flow prediction.
	
	\item \textbf{STDGRL (ours)}: The proposed spatiotemporal prediction network based on spatiotemporal dynamic graph relationships for traffic forecasting in metro stations. Compared with the previous methods, our method does not require a predefined spatial graph on the one hand and can perform long-term metro flow prediction on the other hand.
\end{itemize}

\begin{table}[ht]
	\centering
	\caption {Baselines.}
	\label{tab:chap:table_baselines}
	\scalebox{0.6}{
	\begin{tabular}{c||c|c|c|c}
		\hline \hline \text { Model } & \text { Temporal Relation} & \text { Spatial Relation} & \text { Node Embedding } & \text { ST Fusion } \\
		\hline \text { HA } & \checkmark & & & \\
		\text { SVR } & \checkmark & & & \\
		\text { LSTM } & \checkmark & & & \\
		\text { GRU } & \checkmark & & & \\
		\text { T-GCN } & \checkmark & \checkmark & & \\
		\text { DCRNN } & \checkmark & \checkmark & & \\
		\text { STGCN } & \checkmark & \checkmark & & \\
		\text { AGCRN } & \checkmark & \checkmark & \checkmark & \\
		\text { Graph WaveNet } & \checkmark & \checkmark & \checkmark & \\
		\text { STTN } & \checkmark & \checkmark & \checkmark & \\
		\text { Multi-STGCnet } & \checkmark & \checkmark &  & \checkmark \\
		\text { STDGRL (ours) } & \checkmark & \checkmark & \checkmark & \checkmark \\
		\hline \hline
	\end{tabular}}
\end{table}

\subsection{Overall Performance}
\begin{figure}
	\centering
	\subfigure[MAE]{
		\label{fig:subfig:MAE} %% label for first subfigure
		\includegraphics[width=0.8\columnwidth]{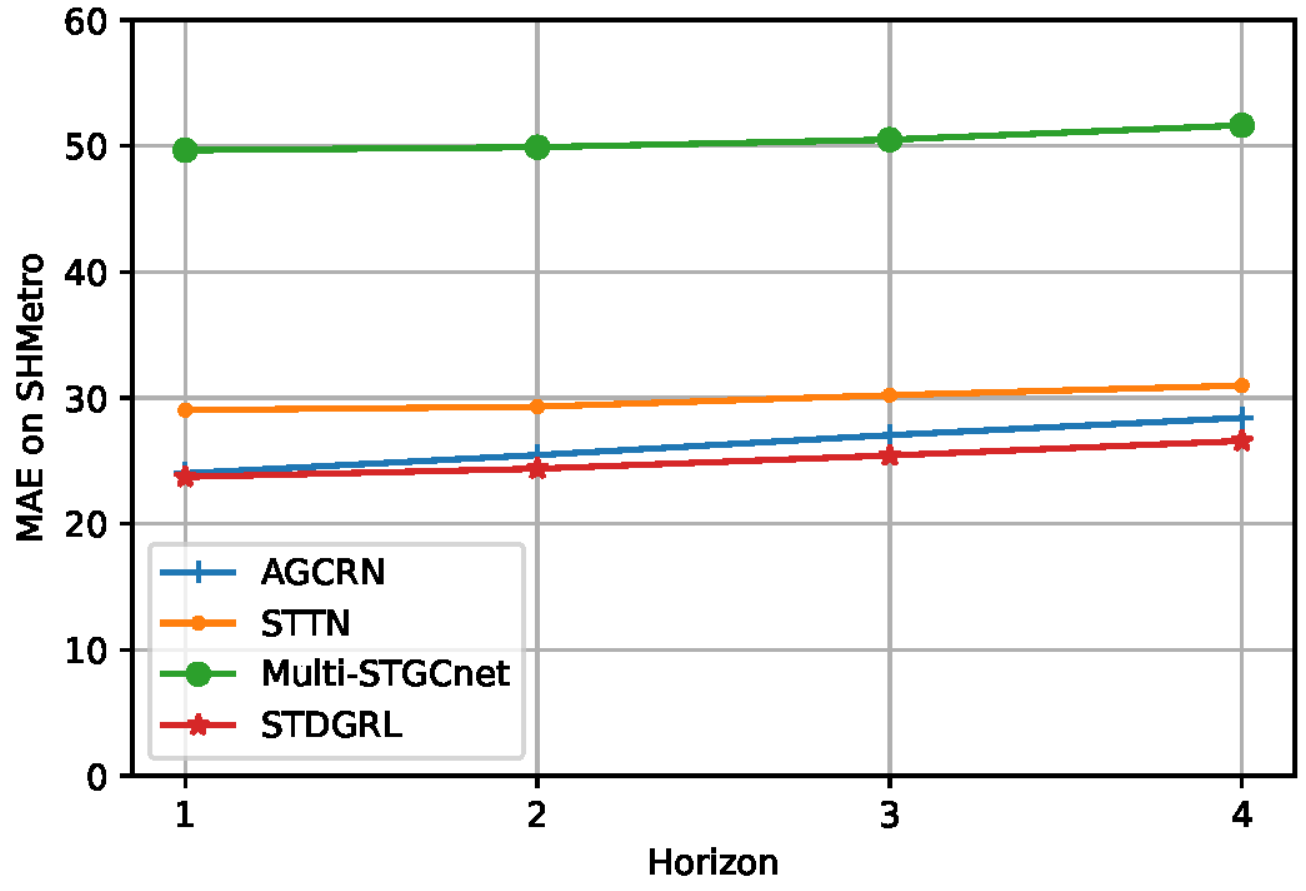}}
	%	\subfigure[RMSE]{
	%		\label{fig:subfig:RMSE} %% label for second subfigure
	%		\includegraphics[width=0.8\columnwidth]{RMSE}}
	\subfigure[MAPE]{
		\label{fig:subfig:MAPE} %% label for third subfigure
		\includegraphics[width=0.8\columnwidth]{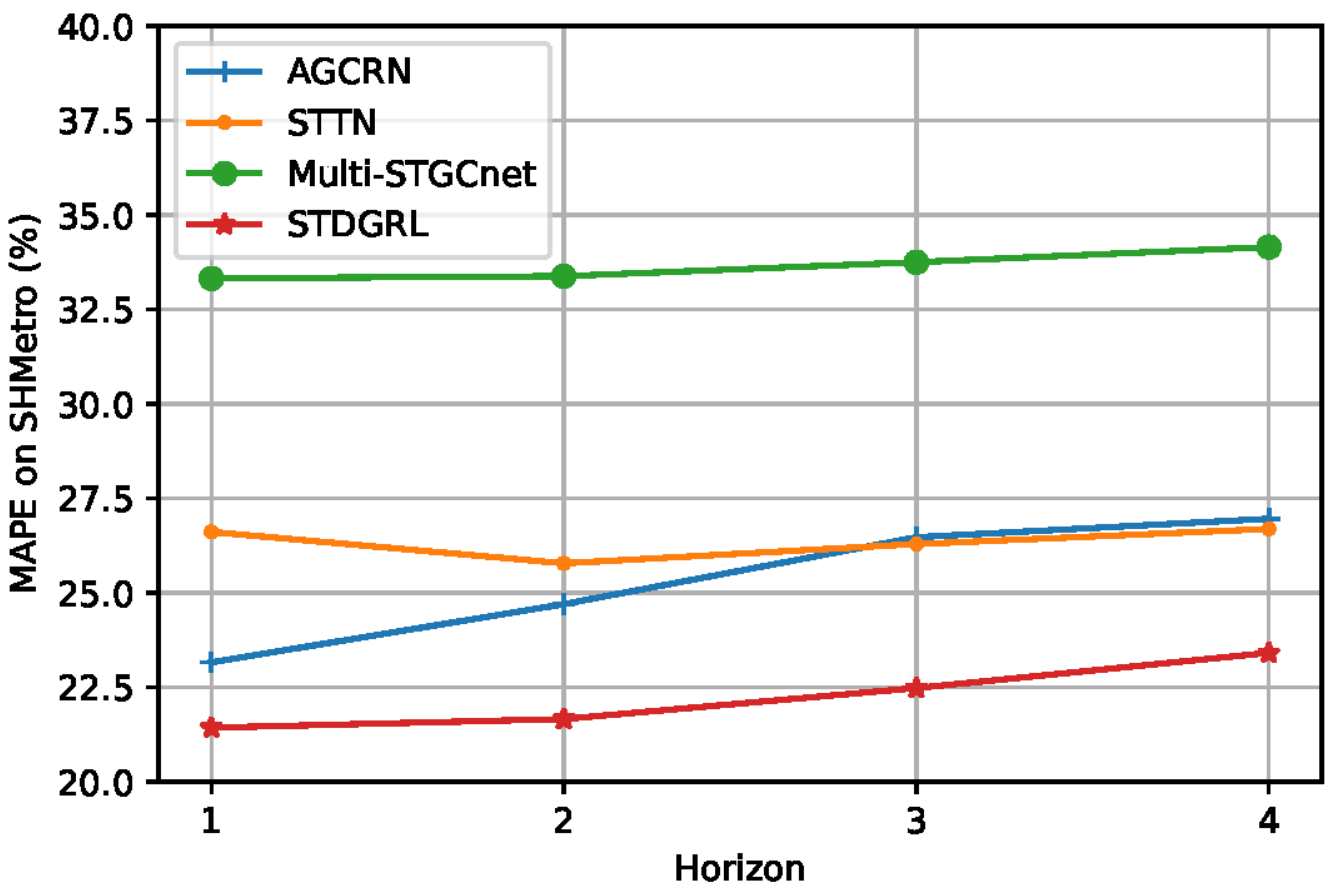}}
	\caption{Prediction performance at each horizon on the SHMetro dataset.}
	\label{fig:subfig} %% label for entire figure
\end{figure}

\begin{figure}[t]
	\centering
	\subfigure[Inflow]{
		\label{fig:subfig:inflow} %% label for first subfigure
		\includegraphics[width=0.8\columnwidth]{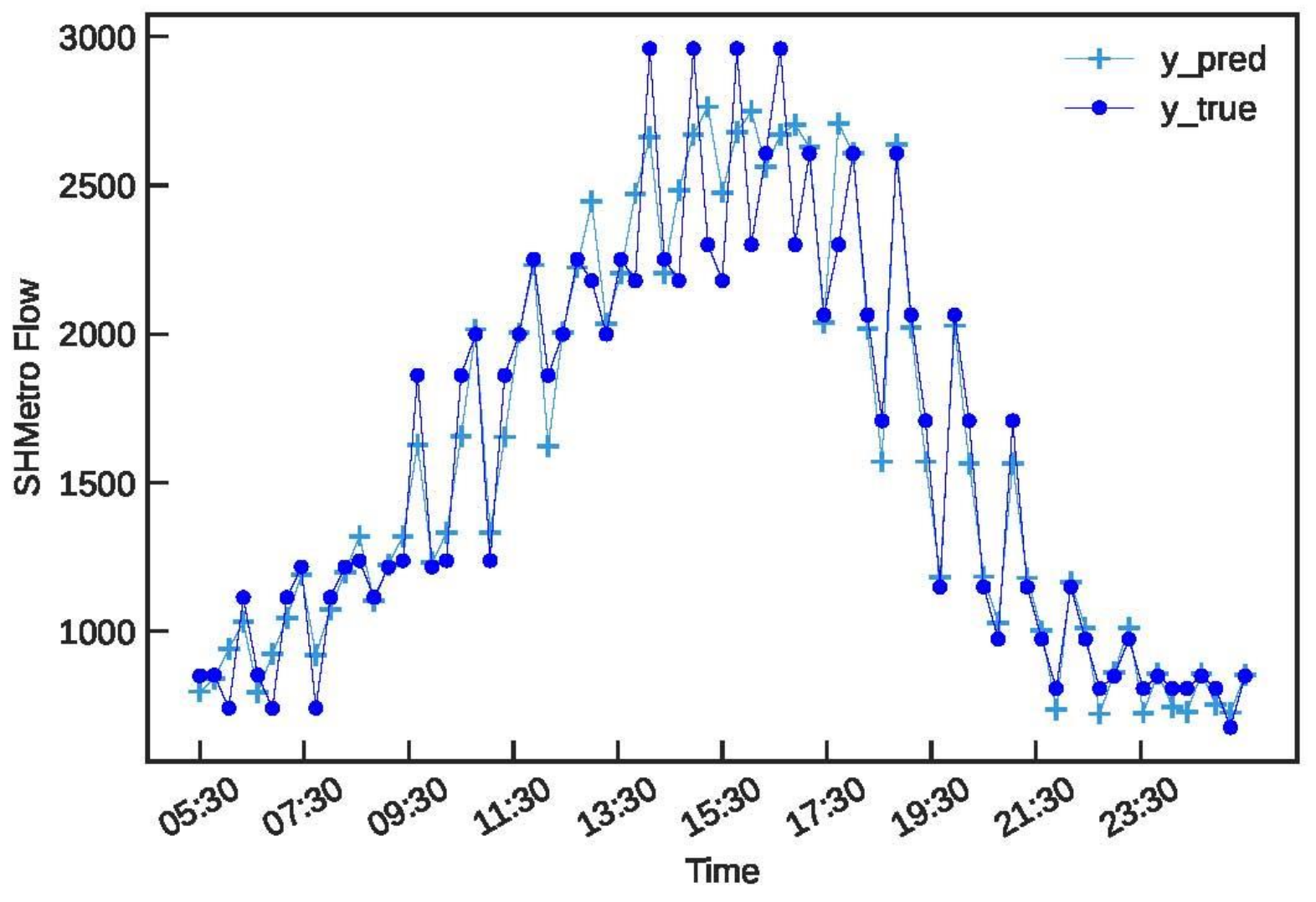}}
	\subfigure[Outflow]{
		\label{fig:subfig:outflow} %% label for second subfigure
		\includegraphics[width=0.8\columnwidth]{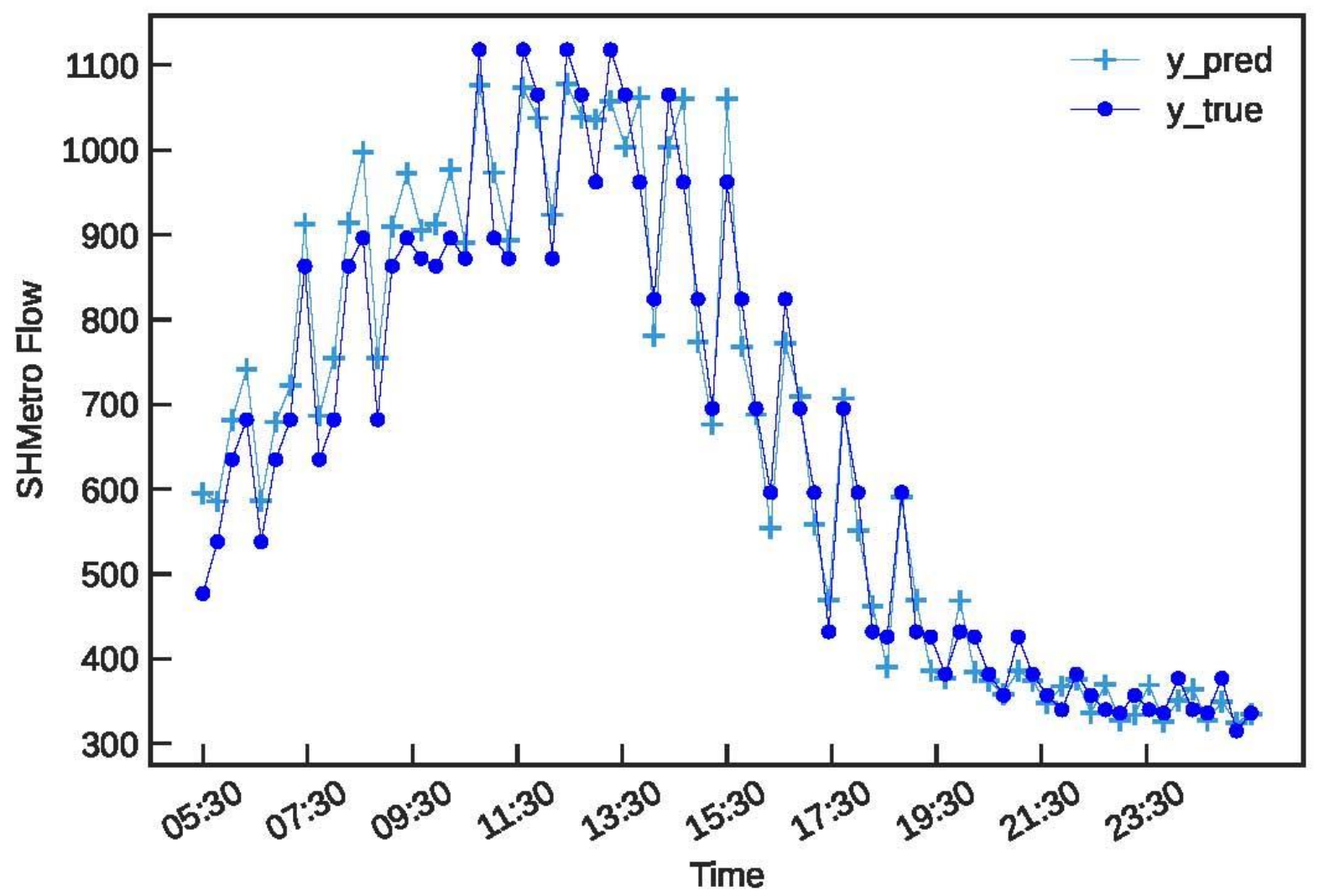}}
	\caption{Inflow and outflow prediction visualization on the SHMetro dataset.}
	\label{fig:subfig_inflow_outflow} %% label for entire figure
\end{figure}

The Table \ref{table_cqmetro} to Table \ref{table_bjmetro} show the overall prediction performance of our method and 11 comparative methods on the Chongqing, Shanghai, Hangzhou and Beijing Metro datasets. In the prediction interval of the next hour, three evaluation indicators MAE, RMSE, and MAPE are used for evaluation. Our proposed STDGRL method has good performance in both short-term and long-term forecasts, as shown in the Figure \ref{fig:subfig}. With the expansion of the prediction interval, the performance of AGCRN method in MAE and MAPE evaluation indicators gradually deteriorates, and the range of change is larger than that of the STDGRL method, indicating that our method has better performance in the long prediction interval. In addition, compared with STTN and Multi-STGCnet methods, our method has obvious performance advantages in both short-term and long-term prediction intervals. We can see that the results of the classical machine learning-based time series forecasting method are not better than the deep learning-based method such as LSTM, GRU methods, indicating that the modeling of no-linear data dependencies in the spatiotemporal data is crucial when making traffic predictions. In addition, we also find that the performance of the traffic prediction models based on graph neural network proposed in recent years are better than LSTM and GRU methods. The reason is that they can capture the spatio-temporal dependence in spatio-temporal graph data better than the deep learning model. We observe that the performance of AGCRN method is better than other baseline models. It significantly improves experimental results, and this method is second only to our method STDGRL. It indicates that the learned spatial relationship from spatiotemporal data can better reflect its spatial dependence. In addition, we also conduct experiments on three public metro datasets, and the experimental results are shown in Table \ref{table_shmetro}, Table \ref{table_hzmetro} and Table \ref{table_bjmetro}. On the Shanghai Metro dataset, the STDGRL still has significant advantages. Figure \ref{fig:subfig_inflow_outflow} shows the inflow and outflow prediction performance at one day in the SHMetro dataset. This dataset contains 288 stations more than other cities stations like Chongqing, Hangzhou. It shows that our proposed method performs well on a small number of stations and also achieves good experimental results on a large number of stations.

\begin{table*}[t]\scriptsize % \tiny, \scriptsize, \footnotesize from small to large
	\centering
	\caption{Performance comparison of baseline methods on CQMetro dataset.}
	\label{table_cqmetro}
	\begin{tabular}{cc|ccc|ccc|ccc|ccc}
		\toprule[1pt]
		
		\multicolumn{2}{c|}{\multirow{2}*{Model}} &\multicolumn{3}{|c}{15min}&\multicolumn{3}{|c}{30min}&\multicolumn{3}{|c}{45min}&\multicolumn{3}{|c}{60min}\\ \cline{3-14}
		
		\multicolumn{2}{c|}{} & MAE  & RMSE & MAPE & MAE & RMSE & MAPE & MAE & RMSE &MAPE & MAE & RMSE &MAPE\\
		\midrule[0.5pt]
		
		\multicolumn{2}{c|}{HA}	&22.6873	&44.4672	&1.0492	&22.6873	&44.4672	&1.0492	&22.6873	&44.4672 &1.0492 &22.6873	&44.4672 &1.0492\\ 
		
		\multicolumn{2}{c|}{SVR}	&23.0547	&47.6177	&1.2276	&23.4774	&48.2486	&1.2468	&24.3611	&49.8825 &1.2916 &25.4671	&51.9743 &1.3365\\

		\multicolumn{2}{c|}{LSTM} &14.9742 &29.9744 &0.8541 &14.8153 &29.3528	&0.9047	&15.0312	&29.7017 &0.9033 &15.4540 &30.0687	&1.1451\\

		\multicolumn{2}{c|}{GRU}	&14.4919	&27.8482	&0.8277	&14.2734	&26.5776	&0.8878	&14.5248	&26.9794 &0.9767 &15.1376 &27.7783	&1.3291 \\

		\multicolumn{2}{c|}{T-GCN}	&21.6926	&35.1871 &1.6189	&23.2144	&37.0886	&1.9629	&25.2631	&41.2690 &2.0706 &27.0057	&45.2837	&2.4766\\

		\multicolumn{2}{c|}{DCRNN}	&15.4474	&29.1529	&0.8449	&16.1849	&29.6531	&0.9011	&17.4521	&32.8911 &0.9376 &18.6397	&36.2879	&1.0071\\ 
		
		\multicolumn{2}{c|}{STGCN}	&15.1819	&27.3737	&0.9369	&15.9876	&27.9732	&1.0453	&17.5419	&31.0785 &1.1385 &19.0031	&34.5544	&1.2687\\ 
		
		\multicolumn{2}{c|}{AGCRN}	&\underline{12.8773}	&\underline{23.5496}	&\underline{0.7516}	&\underline{12.9903}	&\underline{23.0776}	&\underline{0.8022}	&\underline{13.1126}	&\underline{23.4551} &\underline{0.8241} &\underline{13.4228}	&\underline{23.8650}	&0.9641\\ 
		
		\multicolumn{2}{c|}{STTN}	&14.9288	&27.4377	&0.8612	&14.9991	&27.5921	&0.8730	&15.0118	&27.5057 &0.8710 &15.2668	&27.9943	&0.9712\\		
		
		\multicolumn{2}{c|}{Graphwavenet}	&14.1766	&25.1481	&0.7942	&14.3751	&25.1149	&0.8532	&14.9222	&25.9422 &0.9627 &15.4853	&27.0418	&1.1322\\ 
		
		\multicolumn{2}{c|}{Multi-STGCnet}	&18.0469	&40.5857	&0.8304	&17.9263	&39.9126	&0.8821	&17.9012	&39.9775 &0.8321 &17.9893	&40.1645	&\underline{0.8680}\\ 
		
		\multicolumn{2}{c|}{\textbf{STDGRL(ours)}}	&\textbf{12.5579}	&\textbf{22.6911}	&\textbf{0.6878}	&\textbf{12.6113}	&\textbf{22.6256}	&\textbf{0.7521}	&\textbf{12.6616}	&\textbf{22.7524} &\textbf{0.7372} &\textbf{12.8485}	&\textbf{23.0310}	&\textbf{0.8224}\\
		
		\multicolumn{2}{c|}{Improvements}	&+2.48\%	&+3.65\%	&+8.48\%	&+2.92\%	&+1.96\%	&+6.25\%	&+3.44\%	&+3.00\% &+10.54\% &+4.28\%	&+3.49\%	&+5.26\%\\ 
		\bottomrule[1pt]
	\end{tabular}
	%	\label{table_MAP}
\end{table*}

\begin{table*}[t]\scriptsize % \tiny, \scriptsize, \footnotesize from small to large
	\centering
	\caption{Performance comparison of baseline methods on SHMetro dataset.}
	\label{table_shmetro}
	\begin{tabular}{cc|ccc|ccc|ccc|ccc}
		\toprule[1pt]
		
		\multicolumn{2}{c|}{\multirow{2}*{Model}} &\multicolumn{3}{|c}{15min}&\multicolumn{3}{|c}{30min}&\multicolumn{3}{|c}{45min}&\multicolumn{3}{|c}{60min}\\ \cline{3-14}
		
		\multicolumn{2}{c|}{} & MAE  & RMSE & MAPE & MAE & RMSE & MAPE & MAE & RMSE &MAPE & MAE & RMSE &MAPE\\
		\midrule[0.5pt]
		
		\multicolumn{2}{c|}{HA}	& 76.9445 &	169.6002 &	0.9358 &	76.9445 &	169.6002 &	0.9358 & 76.9445 & 169.6002 & 0.9358 & 76.9445 &	169.6002 & 0.9358\\ 
		
		\multicolumn{2}{c|}{SVR} & 89.4518 & 230.2805 &	1.2532 &	91.0132 &	233.0358 &	1.2225 &	94.6976 &	239.6837 & 1.2640 & 100.0826 &	249.1291 & 1.3695\\

		\multicolumn{2}{c|}{LSTM} &48.1613 &108.2152 &0.6381 &53.4732 &125.0903	&0.6604	&57.8482	&136.6724 &0.6826 &64.2742 &156.8241	&0.7489\\

		\multicolumn{2}{c|}{GRU}	&31.2748	&65.8625	&0.3176	&31.6766	&67.4298	&0.3108	&32.5833	&71.2581 &0.3151 &33.7280 &74.3567	&0.3235\\

		\multicolumn{2}{c|}{T-GCN}	&74.6434	&124.6865 &1.3138	&83.4037	&147.2772	&1.3331	&95.1702	&176.0193 &1.5574 &106.0074	&202.7877	&1.8807\\

		\multicolumn{2}{c|}{DCRNN}	&27.9394	&54.2426	&0.2633	&31.9161	&63.9539	&0.2937	&37.2232	&79.1991 &0.3157 &42.0734	&93.8128	&0.3435\\ 
		
		\multicolumn{2}{c|}{STGCN}	&28.2697	&52.2552	&0.3136	&31.8696	&59.3756	&0.3527	&36.9222	&70.1263 &0.4005 &42.0439	&81.2147	&0.4431\\ 
		
		\multicolumn{2}{c|}{AGCRN}	&\underline{24.0087}	&\underline{47.1056}	&\underline{0.2316}	&\underline{25.4590}	&\underline{50.9641}	&\underline{0.2470}	&\underline{27.0434}	&\underline{55.5671} &0.2647 &\underline{28.4134}	&\underline{59.6148}	&0.2696\\ 
		
		\multicolumn{2}{c|}{STTN}	&29.0291	&56.2013	&0.2661	&29.2963	&57.8522	&0.2578	&30.2127	&60.4864 &\underline{0.2629} &30.9729	&60.7344	&\underline{0.2669}\\
		
		\multicolumn{2}{c|}{Graphwavenet}	&26.2299	&50.3182	&0.2448	&28.1380	&54.7953	&0.2689	&30.1868	&59.8046 &0.2868 &32.5230	&65.6138	&0.3265\\ 
		
		\multicolumn{2}{c|}{Multi-STGCnet}	&49.6580	&128.6207	&0.3332	&49.9009	&128.9203	&0.3338	&50.4986	&129.8213 &0.3375 &51.6335	&131.7302	&0.3415\\ 
		
		\multicolumn{2}{c|}{\textbf{STDGRL(ours)}}	&\textbf{23.7239}	&\textbf{46.8692}	&\textbf{0.2143} &\textbf{24.3754} &\textbf{49.2925} &\textbf{0.2166}	&\textbf{25.4230}	&\textbf{52.9028} &\textbf{0.2248}	&\textbf{26.5829}	&\textbf{57.3964} &\textbf{0.2341}\\
		
		\multicolumn{2}{c|}{Improvements}	&+1.19\%	&+0.50\%	&+7.48\%	&+4.26\%	&+3.28\%	&+12.33\%	&+5.99\%	&+4.79\% &+14.49\% &+6.44\%	&+3.72\%	&+12.29\%\\ 
		\bottomrule[1pt]
	\end{tabular}
	%	\label{table_MAP}
\end{table*}

\begin{table*}[t]\scriptsize % \tiny, \scriptsize, \footnotesize from small to large
	\centering
	\caption{Performance comparison of baseline methods on HZMetro dataset.}
	\label{table_hzmetro}
	\begin{tabular}{cc|ccc|ccc|ccc|ccc}
		\toprule[1pt]
		
		\multicolumn{2}{c|}{\multirow{2}*{Model}} &\multicolumn{3}{|c}{15min}&\multicolumn{3}{|c}{30min}&\multicolumn{3}{|c}{45min}&\multicolumn{3}{|c}{60min}\\ \cline{3-14}
		
		\multicolumn{2}{c|}{} & MAE  & RMSE & MAPE & MAE & RMSE & MAPE & MAE & RMSE &MAPE & MAE & RMSE &MAPE\\
		\midrule[0.5pt]
		
		\multicolumn{2}{c|}{HA}	& 71.8148 &	136.8056 &	0.6089 &	71.8148 &	136.8056 &	0.6089 & 71.8148 &	136.8056 & 0.6089 & 71.8148 &	136.8056 & 0.6089\\ 
		
		\multicolumn{2}{c|}{SVR}	&84.8943	&170.9875	&1.7489	&86.6150	&173.4301	&1.7447	&89.1909	&177.8438 &1.7926 &92.4263	&183.5363 &1.8661\\

		\multicolumn{2}{c|}{LSTM} &27.8706 &50.2641 &0.2675 &28.1602 &50.9464	&0.2695	&28.6903	&51.7961 &0.2732 &29.4597 &53.0576	&0.3338\\

		\multicolumn{2}{c|}{GRU}	&27.2826	&48.6847	&0.2523	&27.7143	&49.7454	&\underline{0.2591}	&27.9942	&50.6825 &0.2614 &28.6244 &51.9195	&\underline{0.3024}\\

		\multicolumn{2}{c|}{T-GCN}	&47.3206	&69.9398 &0.7409	&51.0303	&78.8955	&0.7698	&57.6238	&91.5450 &0.8880 &65.0028	&103.6740	&1.2022\\

		\multicolumn{2}{c|}{DCRNN}	&27.1144	&49.5158	&\underline{0.2280}	&31.2308	&58.2314	&0.2616	&36.9020	&70.9692 &0.2855 &42.7503	&85.0528	&0.3243\\ 
		
		\multicolumn{2}{c|}{STGCN}	&28.2432	&49.0484	&0.3032	&32.2267	&56.2076	&0.3548	&37.7572	&65.9376 &0.4239 &44.5799	&77.8010	&0.6117\\ 
		
		\multicolumn{2}{c|}{AGCRN}	&\underline{23.6154}	&\underline{40.3462}	&0.2335	&\underline{24.9422}	&\underline{43.1928}	&0.2647	&\underline{25.9514}	&\underline{45.2841} &0.2544 &\underline{27.4004}	&\underline{46.7793}	&0.3134\\ 
		
		\multicolumn{2}{c|}{STTN}	&28.1227	&48.4724	&0.2408	&28.8057	&49.0463	&0.2753	&28.6228	&49.6005 &\underline{0.2527} &30.6277	&52.4030	&0.3537\\
		
		\multicolumn{2}{c|}{Graphwavenet}	&25.1968	&42.5834	&0.2475	&26.8730	&45.1082	&0.2803	&29.4834	&50.6676 &0.2851 &31.8565	&56.0680	&0.3253\\ 
		
		\multicolumn{2}{c|}{Multi-STGCnet}	&44.4798	&92.4560	&0.3402	&43.7682	&92.1209	&0.3368	&43.8611	&92.6602 &0.3320 &45.1232	&94.0267	&0.3799\\ 
		
		\multicolumn{2}{c|}{\textbf{STDGRL(ours)}}	&\textbf{23.2666}	&\textbf{39.5458}	&\textbf{0.2091} &\textbf{23.7721} &\textbf{40.4317} &\textbf{0.2141}	&\textbf{24.8948}	&\textbf{42.8774} &\textbf{0.2230}	&\textbf{25.8339}	&\textbf{45.1779} &\textbf{0.2570}\\
		
		\multicolumn{2}{c|}{Improvements}	&+1.48\%	&+1.98\%	&+8.28\%	&+4.69\%	&+6.39\%	&+17.40\%	&+4.07\%	&+5.31\% &+11.75\% &+5.72\%	&+3.42\%	&+15.02\%\\ 
		\bottomrule[1pt]
	\end{tabular}
	%	\label{table_MAP}
\end{table*}

\begin{table*}[t]\scriptsize % \tiny, \scriptsize, \footnotesize from small to large
	\centering
	\caption{Performance comparison of baseline methods on BJMetro dataset.}
	\label{table_bjmetro}
	\begin{tabular}{cc|ccc|ccc|ccc|ccc}
		\toprule[1pt]
		
		\multicolumn{2}{c|}{\multirow{2}*{Model}} &\multicolumn{3}{|c}{15min}&\multicolumn{3}{|c}{30min}&\multicolumn{3}{|c}{45min}&\multicolumn{3}{|c}{60min}\\ \cline{3-14}
		
		\multicolumn{2}{c|}{} & MAE  & RMSE & MAPE & MAE & RMSE & MAPE & MAE & RMSE &MAPE & MAE & RMSE &MAPE\\
		\midrule[0.5pt]
		
		\multicolumn{2}{c|}{HA}	&95.7779   &207.2597	  &0.7318	  &95.7779	  &207.2597	  &0.7318	  &95.7779   &207.2597	  &0.7318   &95.7779   &207.2597	  &0.7318  \\ 
		
		\multicolumn{2}{c|}{SVR}	&133.3139 	&313.8002 	&2.1439 	&135.0974 	&317.4431 	&2.1201  &138.4471 	&323.1964  &2.1260  &143.1395 	&330.3005  &2.1365 \\

		\multicolumn{2}{c|}{LSTM} &99.2410  &243.2237  &1.9165  &102.5640  &245.3053 	&2.0540 	&107.7217 	&248.8519  &2.5461  &115.4902  &257.5999 	&3.8183 \\

		\multicolumn{2}{c|}{GRU}	&96.3814 	&237.3694 	&1.7907 	&96.5315 	&238.4541 	&1.8164 	&98.0139 	&240.9763  &2.0611  &101.0722  &245.6744 	&3.3837 \\

		\multicolumn{2}{c|}{T-GCN}	&97.1880 	&157.4604  &1.8642 	&110.1468 	&183.8415 	&2.2288 	&126.7785 	&217.8278  &3.1665  &141.9155 	&250.9208 	&4.6435 \\

		\multicolumn{2}{c|}{DCRNN}	&32.4452 	&67.2273 	&0.2861 	&38.5430 	&81.8017 	&0.3725 	&47.0715 	&103.1199  &0.5501  &55.3968	&125.2164 	&0.8658 \\ 
		
		\multicolumn{2}{c|}{STGCN}	&32.1576 	&62.6209 	&0.3366 	&37.8507 	&71.9395 	&0.4629 	&44.9624 	&84.5158  &0.7980  &50.8894 	&96.7363 	&1.5807 \\ 
		
		\multicolumn{2}{c|}{AGCRN}	&\underline{25.1688} 	&\underline{47.8686}	&\underline{0.2397} 	&\underline{25.3167} 	&\underline{47.2164} 	&\underline{0.2669} 	&\underline{26.2948} 	&\underline{48.9524}  &\underline{0.3599}  &\underline{26.9285} 	&\underline{50.8812} 	&\underline{0.5362} \\ 
		
		\multicolumn{2}{c|}{STTN}	&35.6133	&78.4165	&0.3647	&32.7436	&63.3843	&0.3284	&33.2021	&62.4016 &0.4469 &35.8133	&68.6054	&0.9178\\
		
		\multicolumn{2}{c|}{Graphwavenet}	&30.0961 	&54.7262 	&0.3078 	&32.2696 	&59.0870 	&0.3418 	&34.8733 	&64.4616  &0.4582  &37.7106 	&70.6784 	&0.8562 \\ 
		
		\multicolumn{2}{c|}{Multi-STGCnet}	&74.9387 	&205.3702	&0.7335 	&74.8064 	&205.1637 	&0.7601 	&75.0618 	&205.4398  &0.9342  &75.5030 	&206.6044 	&1.2342 \\ 
		
		\multicolumn{2}{c|}{\textbf{STDGRL(ours)}}	&\textbf{21.8468}	&\textbf{41.2336}	&\textbf{0.2015} &\textbf{22.3419} &\textbf{42.3507} &\textbf{0.2167}	&\textbf{22.8053}	&\textbf{43.1799} &\textbf{0.2908}	&\textbf{22.8942}	&\textbf{43.3726} &\textbf{0.4393}\\
		
		\multicolumn{2}{c|}{Improvements}	&+13.20\% 	&+13.86\% 	&+15.94\%  	&+11.75\%  	&+10.31\%  	&+18.80\%  	&+13.27\%  	&+11.79\%   &+19.21\%   &+14.98\%  	&+14.76\%  	&+18.07\%\\ 
		\bottomrule[1pt]
	\end{tabular}
	%	\label{table_MAP}
\end{table*}

In summary, the experiment result demonstrates that STDGRL can learn the spatial and temporal relation from the metro spatio-temporal graph of different scales and achieve promising predictions performance.

\begin{table*}[t]\scriptsize % \tiny, \scriptsize, \footnotesize from small to large
	\centering
	\caption{Analysis of ablation study on CQMetro dataset.}
	\label{table_ablation_cqmetro}
	\begin{tabular}{cc|ccc|ccc|ccc|ccc}
		\toprule[1pt]
		
		\multicolumn{2}{c|}{\multirow{2}*{Model}} &\multicolumn{3}{|c}{15min}&\multicolumn{3}{|c}{30min}&\multicolumn{3}{|c}{45min}&\multicolumn{3}{|c}{60min}\\ \cline{3-14}
		
		\multicolumn{2}{c|}{} & MAE  & RMSE & MAPE & MAE & RMSE & MAPE & MAE & RMSE &MAPE & MAE & RMSE &MAPE\\
		\midrule[0.5pt]
		
		\multicolumn{2}{c|}{GCGRU(T-GCN)}	&21.6926	&35.1871	&1.6189	&23.2144	&37.0886	&1.9629	&25.2631	&41.2690 &2.0706 &27.0057	&45.2837 &2.4766\\ 
		
		\multicolumn{2}{c|}{STDGRL-NAPL}	&13.1137	&24.0971	&0.7766	&13.2085	&23.5815	&0.8447	&13.5121	&24.0103 &0.9089 &13.8917	&24.6763 &1.0789\\

		\multicolumn{2}{c|}{STDGRL-Transformer} &\textbf{12.7363} &23.1410 &0.7404 &\textbf{12.7988} &\textbf{22.8028}	&0.8151	&\textbf{12.8622}	&\textbf{22.7317} &0.8111 &\textbf{12.9504} &\textbf{22.9711}	&\textbf{0.8411}\\

		\multicolumn{2}{c|}{STDGRL-GRU-Transformer}	&12.8098	&\textbf{22.9895}	&\textbf{0.7215}	&12.8164	&22.8653	&\textbf{0.7724}	&12.9719	&23.1255 &\textbf{0.7784} &13.0303 &23.2679	&0.8636 \\ 
		
		\bottomrule[1pt]
	\end{tabular}
\end{table*}

\begin{table*}[t]\scriptsize % \tiny, \scriptsize, \footnotesize from small to large
	\centering
	\caption{Analysis of ablation study on SHMetro dataset.}
	\label{table_ablation_shmetro}
	\begin{tabular}{cc|ccc|ccc|ccc|ccc}
		\toprule[1pt]
		
		\multicolumn{2}{c|}{\multirow{2}*{Model}} &\multicolumn{3}{|c}{15min}&\multicolumn{3}{|c}{30min}&\multicolumn{3}{|c}{45min}&\multicolumn{3}{|c}{60min}\\ \cline{3-14}
		
		\multicolumn{2}{c|}{} & MAE  & RMSE & MAPE & MAE & RMSE & MAPE & MAE & RMSE &MAPE & MAE & RMSE &MAPE\\
		\midrule[0.5pt]
		
		\multicolumn{2}{c|}{GCGRU(T-GCN)}	&74.6434  & 124.6865  & 1.3138  & 83.4037  & 147.2772  & 1.3331  & 95.1702  & 176.0193  & 1.5574  & 106.0074  & 202.7877  & 1.8807\\

		\multicolumn{2}{c|}{STDGRL-Transformer} & 24.6472  & 47.9299  & 0.2297  & 25.5251  & 50.9482  & 0.2358  & \textbf{26.7347}  & 55.5103  & 0.2474  & \textbf{28.0522}  & 61.0607  & 0.2565\\

		\multicolumn{2}{c|}{STDGRL-GRU-Transformer}	& 24.5923  & \textbf{47.8691}  & \textbf{0.2191}  & \textbf{25.4948}  & \textbf{50.5436}  & \textbf{0.2219}  & 26.9211  & \textbf{54.8073}  & \textbf{0.2336}  & 28.3289  & 59.8896  & \textbf{0.2431} \\ 
		
		\multicolumn{2}{c|}{STDGRL-NAPL}	&\textbf{24.5406}  & 48.5099  & 0.2415  & 26.2928  & 53.3123  & 0.2548  & 27.7053  & 56.4908  & 0.2744  & 28.9449  & \textbf{59.2060}  & 0.2860\\ 
		
		\bottomrule[1pt]
	\end{tabular}
\end{table*}

\begin{table*}[t]\scriptsize % \tiny, \scriptsize, \footnotesize from small to large
	\centering
	\caption{Analysis of ablation study on HZMetro dataset.}
	\label{table_ablation_hzmetro}
	\begin{tabular}{cc|ccc|ccc|ccc|ccc}
		\toprule[1pt]
		
		\multicolumn{2}{c|}{\multirow{2}*{Model}} &\multicolumn{3}{|c}{15min}&\multicolumn{3}{|c}{30min}&\multicolumn{3}{|c}{45min}&\multicolumn{3}{|c}{60min}\\ \cline{3-14}
		
		\multicolumn{2}{c|}{} & MAE  & RMSE & MAPE & MAE & RMSE & MAPE & MAE & RMSE &MAPE & MAE & RMSE &MAPE\\
		\midrule[0.5pt]
		
		\multicolumn{2}{c|}{GCGRU(T-GCN)}	&47.3206  & 69.9398  & 0.7409  & 51.0303  & 78.8955  & 0.7698  & 57.6238  & 91.5450  & 0.8880  & 65.0028  & 103.6740  & 1.2022\\ 
		
		\multicolumn{2}{c|}{STDGRL-GRU-Transformer}	&24.2082  & 40.6312  & 0.2244  & 25.2125  & 42.4502  & 0.2430  & 26.9467  & 45.9737  & 0.2595  & 29.4166  & 50.9448  & 0.3488 \\ 
		
		\multicolumn{2}{c|}{STDGRL-NAPL}	&23.9697  & 41.8146  & \textbf{0.2143}  & 25.6265  & 44.7244  & 0.2400  & 27.0741  & 47.1538  & 0.2456  & 28.8701  & 49.7926  & \textbf{0.2871}\\ 
		
		\multicolumn{2}{c|}{STDGRL-Transformer} &\textbf{23.2615}  & \textbf{39.7872}  & 0.2178  & \textbf{24.0419}  & \textbf{40.9422}  & \textbf{0.2300}  & \textbf{24.8309}  & \textbf{42.4351}  & \textbf{0.2415}  & \textbf{26.0904}  & \textbf{45.1816}  & 0.2977\\ 
		
		\bottomrule[1pt]
	\end{tabular}
\end{table*}

\begin{table*}[t]\scriptsize % \tiny, \scriptsize, \footnotesize from small to large
	\centering
	\caption{Analysis of ablation study on BJMetro dataset.}
	\label{table_ablation_bjmetro}
	\begin{tabular}{cc|ccc|ccc|ccc|ccc}
		\toprule[1pt]
		
		\multicolumn{2}{c|}{\multirow{2}*{Model}} &\multicolumn{3}{|c}{15min}&\multicolumn{3}{|c}{30min}&\multicolumn{3}{|c}{45min}&\multicolumn{3}{|c}{60min}\\ \cline{3-14}
		
		\multicolumn{2}{c|}{} & MAE  & RMSE & MAPE & MAE & RMSE & MAPE & MAE & RMSE &MAPE & MAE & RMSE &MAPE\\
		\midrule[0.5pt]
		
		\multicolumn{2}{c|}{GCGRU(T-GCN)}	&97.1880  & 157.4604  & 1.8642  & 110.1468  & 183.8415  & 2.2288  & 126.7785  & 217.8278  & 3.1665  & 141.9155  & 250.9208  & 4.6435\\ 
		
		\multicolumn{2}{c|}{STDGRL-NAPL}	&26.2780  & 50.5006  & 0.2732  & 26.8332  & 50.8173  & 0.2990  & 28.3084  & 54.1128  & 0.4353  & 29.2143  & 56.2724  & 0.8812\\ 
		
		\multicolumn{2}{c|}{STDGRL-Transformer} &24.0629  & 44.3345  & \textbf{0.2393}  & \textbf{24.6776}  & 45.9240  & \textbf{0.2621}  & \textbf{25.6731}  & 48.4556  & \textbf{0.3524}  & \textbf{26.0899}  & 49.2110  & \textbf{0.6429}\\ 
		
		\multicolumn{2}{c|}{STDGRL-GRU-Transformer}	&\textbf{23.9464}  & \textbf{43.5017}  & 0.2435  & 24.7591  & \textbf{45.8199}  & 0.2759  & 25.9610  & \textbf{48.3604}  & 0.4270  & 26.1943  & \textbf{48.8126}  & 0.8663 \\ 
		
		\bottomrule[1pt]
	\end{tabular}
\end{table*}

\begin{figure}[t]
	\centering
	\subfigure[MAE]{
		\label{fig:subfig:ablation_MAE} %% label for first subfigure
		\includegraphics[width=0.8\columnwidth]{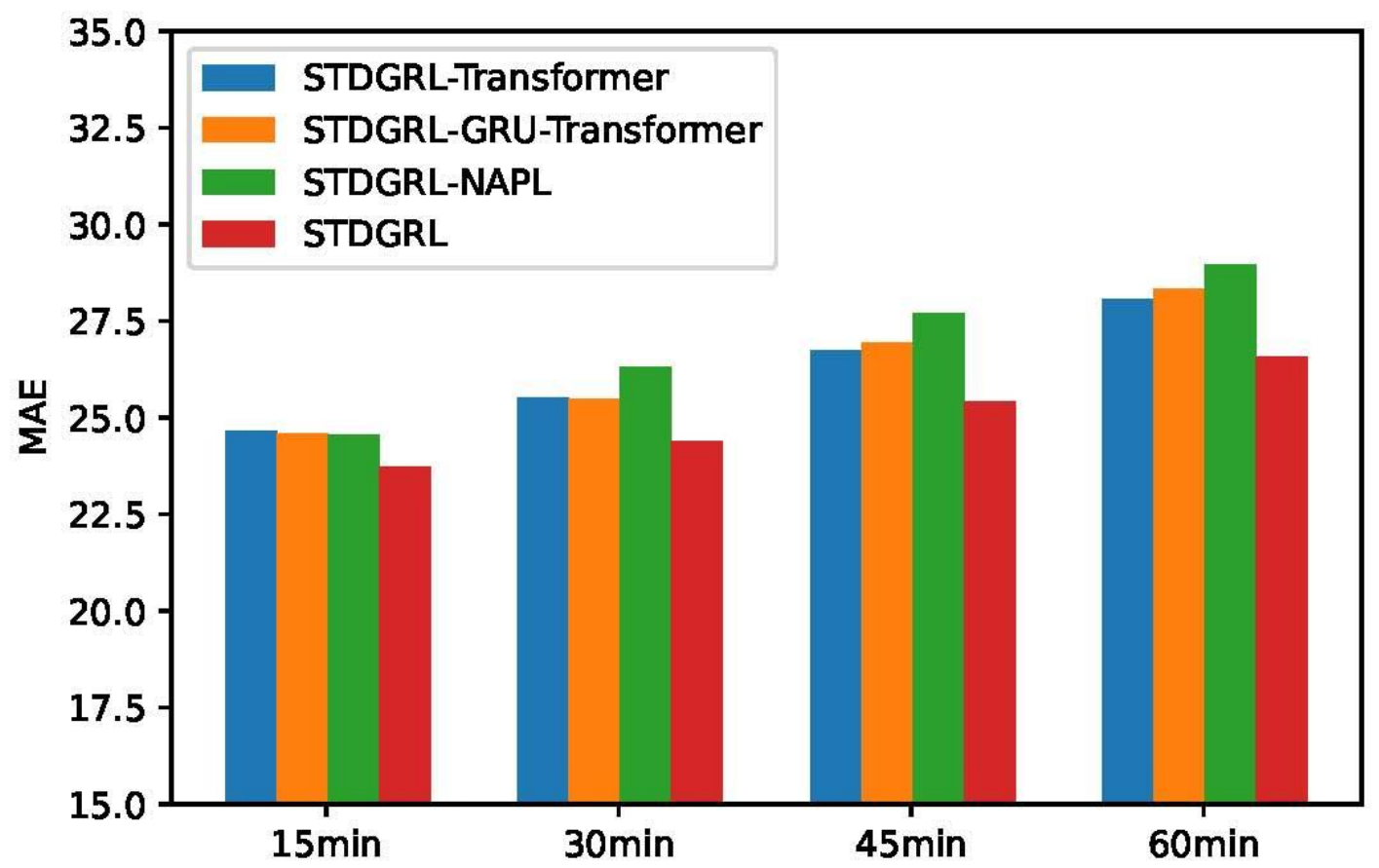}}
	\subfigure[MAPE]{
		\label{fig:subfig:ablation_MAPE} %% label for second subfigure
		\includegraphics[width=0.8\columnwidth]{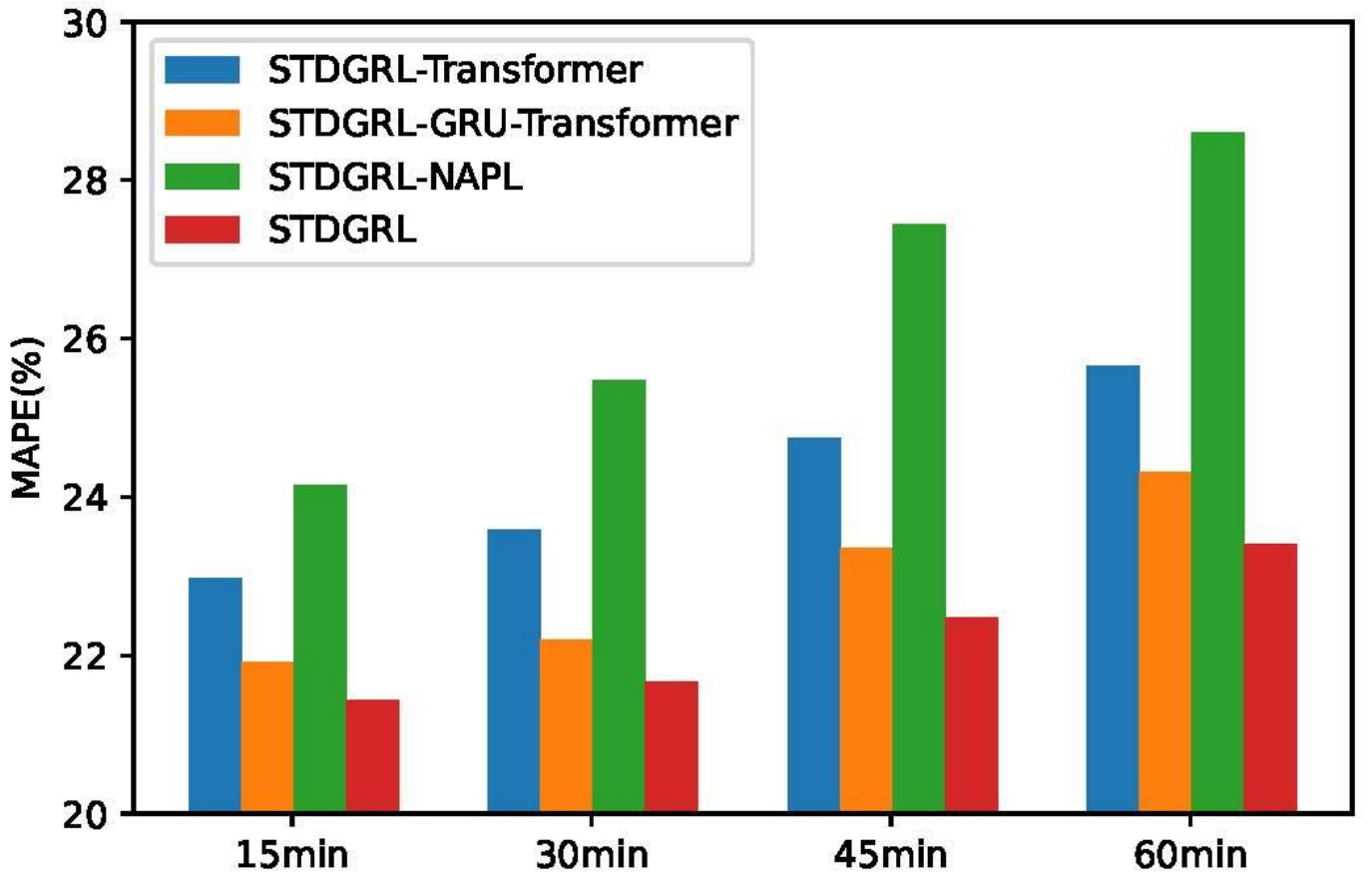}}
	\caption{Ablation study performance on the SHMetro dataset.}
	\label{fig:subfig_ablation} %% label for entire figure
\end{figure}

\subsection{Ablation Study}
We design a comprehensive ablation study to evaluate the performance of STDGRL. The baseline model of our ablation study is GCGRU(T-GCN). This model is a classical traffic forcasting method, which combine GCN and GRU for capturing spatio-temporal dependencies. And we remove the NAPL component from the STDGRL model to construct STDGRL-NAPL. STDGRL-Transformer and STDGRL-GRU-Transformer are the variants of our STDGRL respectively, which remove GRU module, GRU and Transformer module from STDGRL model. The experimental result on the four datasets are illustrated in Table \ref{table_ablation_cqmetro}, Table \ref{table_ablation_shmetro}, Table \ref{table_ablation_hzmetro} and Table \ref{table_ablation_bjmetro}. And we also show the ablation study performance on the SHMetro dataset in Figure \ref{fig:subfig_ablation}. We can observe that: 1) The results in the Table show that the performance of GCGRU (T-GCN) is not as good as that of the other three comparison models, which may be due to its use of pre-defined graphs and difficulty in capturing complex spatial dependencies between nodes. 2) Compared with the STDGRL model, the performance of the STDGR-NAPL model decreases by a large proportion and is inferior to STDGR-Transformer and STDGR-GRU-Transformer, indicating that it is necessary to capture node-specific traffic patterns in the STDGRL model. 3) After Transformer and GRU modules are removed from the STDGRL model, the performance is lower than that of the STDGRL model, but better than that of the STDGRL-NAPL model, indicating the necessity of using short-term and long-term time series prediction modules in the STDGRL model. And it also demonstrates learning the specific traffic patterns of nodes is more important than learning temporal dependencies.

Overall, our NAPL, DSRL and temporal learning modules jointly boost the prediction performance of the STDGRL model.

\section{Conclusion} \label{section: conclusion}
We proposed STDGRL, a novel spatio-temporal dynamic graph relationship learning model, for predicting multi-step passenger inflow and outflow in urban metro stations. STDGRL can capture the traffic patterns of different metro stations and the dynamic spatial dependencies between metro stations. In addition, STDGRL can capture long-term temporal relationship dependencies for long-term metro flow prediction. We validated our model on real metro datasets in 4 cities and experimental results achieved significant performance improvements over 11 baselines. 

In future work, we plan to research the influence of weather, events and POI on the change of metro passenger flow, and the detection and prediction of sudden large passenger flow in metro stations.

\section*{Acknowledgment}
This research was supported by the National Key R\&D Program of China (2019YFB2101801) and the National Natural Science Foundation of China (No. 62176221).

\ifCLASSOPTIONcaptionsoff
  \newpage
\fi



%\bibliographystyle{IEEEtran}
\vfill
\bibliographystyle{IEEEtran}
%\bibliography{reference.bib}
\bibliography{reference}

\end{document}